\documentclass[conference,letterpaper]{IEEEtran}
\usepackage{fancyhdr}
\usepackage{fancyhdr}
\usepackage{enumerate}
\usepackage{cite}      % Written by Donald Arseneau
                        
\usepackage{graphicx}  % Written by David Carlisle and Sebastian Rahtz

\usepackage{amsmath}   % From the American Mathematical Society
\begin{document}
\title{ Ship Detection and Segmentation using Image Correlation}
\IEEEoverridecommandlockouts
\IEEEpubid{\makebox[\columnwidth]{978-1-4799-0652-9/13/\$31.00~\copyright2013
IEEE \hfill} \hspace{\columnsep}\makebox[\columnwidth]{ }}

\author{\IEEEauthorblockN{Alexander Kadyrov, Hui Yu and Honghai Liu}
\IEEEauthorblockA{School of Creative Technologies,
University of Portsmouth, 
Portsmouth, UK}}
%\maketitle
\thispagestyle{plain}
\fancypagestyle{plain}{
\fancyhf{}	% clear all header and footer fields
\fancyfoot[L]{}
\fancyfoot[C]{}
\fancyfoot[R]{}
\renewcommand{\headrulewidth}{0pt}
\renewcommand{\footrulewidth}{0pt}
}
\pagestyle{fancy}{
\fancyhf{}
\fancyfoot[R]{}}
\renewcommand{\headrulewidth}{0pt}
\renewcommand{\footrulewidth}{0pt}
\maketitle
\begin{abstract}
There have been intensive research interests in ship detection and segmentation due to high demands on a wide range of civil applications in the last two decades. However, existing approaches, which are mainly based on statistical properties of images, fail to detect smaller ships and boats. Specifically, known techniques are not robust enough in view of inevitable small geometric and photometric changes in images consisting of ships. In this paper a novel approach for ship detection is proposed based on correlation of maritime images. The idea comes from the observation that a fine pattern of the sea surface changes considerably from time to time whereas the ship appearance basically keeps unchanged. We want to examine whether the images have a common unaltered part, a ship in this case. To this end, we developed a method - Focused Correlation (FC) to achieve robustness to geometric distortions of the image content. Various experiments have been conducted to evaluate the effectiveness of the proposed approach.
\end{abstract}

\IEEEpeerreviewmaketitle

%%%%%%%%%%%%%%%%%%%%%%%%%%%%%%%%%%%%%%%%%%%%%%%==================
%\maketitle


% Will be ignored for other modes.
\IEEEpeerreviewmaketitle

\section{Introduction}

Robustly detecting ships plays a crucial role in civil applications such as drug-smuggling ships detection. Ship detection problems have been researched intensively. A review \cite{OLSEN} on this topic includes up to 500 literature entries. Generally speaking, two types of techniques have been used for ship detection \cite{corbane2010complete}.  The most popular one is the synthetic aperture radar (SAR) technique, which is fairly robust to various weather conditions. It is based on an invasive technology - the scene is illuminated with radio rays. The detection is achieved through a  series of processing of reflected signals. A side-effect of this technique is that the airborne surveillance system reveals itself. As a result, people on the detected ship may be aware that they are under surveillance. For smaller ships and boats SAR technique is less efficient  \cite{ herselman2008analysis}, and thus remains an open problem which is commonly interpreted as the need for ``accurate empirical modelling of sea'' to separate the boat from the sea.

An alternative, either optical, or visual based detection is lesser developed  \cite{zhu2010novel, corbane2010complete}, though it is considered to be important. For example, in \cite{House2012} it is stated that UK will have ``a range of maritime surveillance resources available in 2020, operating in the audio, visual and electronic spectra''. It should be non-invasive and does not require special equipments. Then it is preferable for the purpose of non-invasive detection of small vessels by unmanned aerial vehicle (UAV), and therefore it is appropriate for improving maritime border surveillance of small ships and boats in order to detect illegal activities such as drug and human trafficking and illegal fishing activities etc.

To the best of our knowledge, most ship detection methods currently operate with one image, and apply thresholding that follows a preprocessing procedure \cite{purohit2008target}. The rationality behind this is straightforward: an experienced human operator is able to distinguish a ship in the surrounding sea based on the fact that a ship has a specific color and shape, and the sea surface has a particular texture. Computers can rely upon the same assumptions. Hence those standard approaches are mainly based on a variety of segmentation techniques and shape analysis methods for suspected inclusions \cite{bingjie2011ship, yu2012automated, sun2012new, bivisual,xiangwei2012fast}, where most advanced techniques model sea patterns by specially designed random fields, and they also model a ship as an elongated inclusion.

\IEEEpubidadjcol
Here in this paper we propose a completely different paradigm. Instead of analysing a single picture, a pair of pictures is considered. The method is based on the observation as follows. During a short time $t_\text{sea}$ the wave pattern changes and hence the water area cannot match with its previous state. On the contrary, a ship's shape does not change much within this time interval, and she can be merely displaced and/or undergo a small rotation in the image, due to movement of the ship and the aerial based camera. Hence, the two images of the sea do not correlate, whereas the two images of the same boat do correlate. This observation forms the principle of the presented research in this paper. We correlate the two images, and if the correlation value is significant, it is concluded that a boat is present in the image. However, in practice a problem arises due to such changes in the ship appearance that the correlation algorithm overlooks the ship presence. To overcome this difficulty, we re-interpret the visual information by creating a controlled uncertainty to combat possible changed in the ship appearance in the video sequence. %In this paper we shall describe the technique, but omit its theoretical argumentation. Instead, we present a heuristic explanation and then an experimental validation.    

The reminder of this paper is organised as follows: Section \ref{Task} presents an explanation of the task and scenarios of ship detection; Section \ref{Corr} gives an overview of available correlation methods, shows a need of their improvement, and suggests such an improvement;  
Section \ref{separ1} explains how a ship and sea can be separated, and the presence of the ship can be detected.  However, the obtained there information is not equivalent to segmentation of the ship. In the Section \ref{compar} we compare different methods for ship detection.  Section \ref{ImSepar} eventually provides segmentation of the detected ship.
Finally,  we conclude this paper with discussion in Section \ref{discussion}.

\section{Task and problem formulation} \label{Task}
Imagine an unmanned aeriall vehicle (UAV) hovering over the sea as a part of a maritime  surveillance system. An on-board camera takes images and sends them to the base for human operators. In this scenario two problems arise. Firstly, most of images contain inadequate information, majority showing an empty sea, or repeatedly sending the same information back to the base already observed previously. Secondly, the human operator is overloaded with a bulk of redundant information and thus can make mistakes when human attention is eventually required to analyse a non-standard situation. An ideal solution to these problems would be to enable on-board computer to automatically analyse images, and send out only those images that do require human attention. Moreover, an on-board computer can take a part in controlling the flight of the UAV \cite{tang2012evolutionary}, taking more pictures of a discovered boat from different perspective, and send them to the base. One of the most important steps of this procedure is an ability to detect and delineate ships and boats, because only images containing boats and ships can raise further interest. We present our work in this direction. 

To narrow the task and to specify the visual information of interest, we point to the following advantages that an UAV can provide. The first advantage is a high definition camera and a powerful computer.  This means that we can rely upon high resolution  images. Another advantage is a highly accurate positioning system employed by UAV.  With this information we can estimate position of the ship/boat in the image. If a boat is present in one image, then it is assumed to present in another image taken within few seconds. On the other hand, we also know that if the fist image does not contain the boat, then the second image also does not contain it (except the fringe of the image). Assumed high resolution allows to proceed images by parts, where boats will have a better representation in their sizes. Thus we formulate our task as detecting whether a boat is present on the both images taken with a time interval up to a threshold of a few seconds.   

The two images of boats in the described scenarios are shifted and geometrically distorted comparatively to one another, a method is sought that is robust in case of small geometric distortions. In spite of their variety, existing correlation methods in Section \ref{Corr} share the same characteristics concerning geometric distortion of images. A new modification is required to meet the needs of our task.  

\section{Correlation methods}\label{Corr}
\subsection{Standard methods}
To examine whether two images $f$ and $g$ have a common area, that is a boat in our case, a few standard options are available. The general procedure is described as follows. The common area may be shifted from the its original position, so then one has to try all possible shift vectors, displacing the first image, and compare such displaced image with the second image. The comparison includes computation of a similarity or dissimilarity measure. Note that this is not excessively time consuming since the computation uses FFT algorithm. To specify the process, we denote the shift vector by $s$  and denote by $x$ a variable pixel in the picture. The measure depend on $s$ only, so we obtain a function $S(s)$ which is called a matching surface \cite{fitch2005fast}. The usage of the matching surface is as follows. The position of maximum (or minimum in case of dissimilarity) of the matching surface $S$ gives the sought shift vector between the images. If the maximum is not high enough, then it indicates that there is no common area in the two images. 

We consider standard ways of defining the matching surface.   
The first formula is a regular cross-correlation,
\begin{equation} S_0(s) = \sum_x f(x-s)g(x),        \label{S0} \end{equation}
and this can be used as a similarity measure. It can be argued that it is not the best way to compare images, so then we consider its standard alternatives.   

Instead of the initial images, one can firstly correlate modified images, then a more general approach would be to define an operation $\mathcal{O}$ that transforms images to new functions $f_1=\mathcal{O}(f)$ and $g_1=\mathcal{O}(g)$ and correlate
them instead of the initial images. Thus, the generalisation of (\ref{S0}) is the formula 
\begin{equation} S_{\mbox{general}}(s) = \sum_x f_1(x-s)g_1(x)  .      \label{S} \end{equation}
There are three well known ways to define the operation $\mathcal{O}$ in this context. They are
\begin{itemize}
\item  ``Orientation correlation'' \cite{fitch2002orientation}. The operation $\mathcal{O}$ is
taking gradient of the image at each pixel and then normalize it, that is
$\mathcal{O}(f) = \nabla (f)/ \| \nabla (f) \|$, then (\ref{S}), where multiplication means scalar product of vectors, represents orientation correlation. 
We will refer to this particular $\mathcal{O}$ as to an ``orientation operator". 
\item  ``Phase correlation'' \cite{Kuglin1975}.  The operation $\mathcal{O}$ is
retaining only phase information in the image, and ignoring the amplitude information in the frequency domain, that is
$\mathcal{O}(f) =  \mathbf{F}(f)/ | \mathbf{F}(f) |$, (where $\mathbf{F}(f)$ is Fourier transform of $f$) and then (\ref{S}) represents phase correlation. For better results one will need to take care of the image borders, this is discussed in \cite{moisan2011periodic}.  
\item  ``Normalized correlation'' \cite{lewis1995fast}. To define it, a size $n$ of a small sliding window should be chosen.  The operation $\mathcal{O}$ is defined as 
$\mathcal{O}(f) =  (f - m(f,x,n))/\sigma(f,x,n)$, where $m(f,x,L)$ is the mean value of the function $f$ in the sliding square window with side $n$ and with its center at $x$, and  $\sigma(f,x,L)$ is the standard deviation of all the values of this function in the square. 
\end{itemize}
The image $\mathcal{O}(f)$ looks 
more random than the initial $f$ (and this can be proved by statistical tests), so the operation $\mathcal{O}$ randomises the underlying image, for details see 
\cite{Pearson1977,Foroosh2002,Gluckman2005,Zeidler1992}.
Other standard alternatives to formula (\ref{S0}) follow: a dissimilarity measures
\[ S_1(s) = \sum_x (f(x-s)-g(x))^2,        \]
\[ S_2(s) = \sum_x |f(x-s)-g(x)|,     \]
and even a general measure
\[ S_3(s) = \sum_x H(f(x-s)-g(x)),     \]
where $H$ is a loss function associated with robust statistics. For the sake of generalization, the images $f$ and $g$ here also can be substituted by $f_1$ and $g_1$.

Let us analyse these surveyed methods of correlating images. Firstly, $S_1(s)$ can be reduced to $S_0(s)$ in (\ref{S0}). For this one can open the parentheses, and get
$S_1(s) =  \mbox{Const}-2 S_0(s)$,
where $\mbox{Const} = \sum_x f(x-s)^2+\sum_x g(x)^2$ is independent of $s$.

Secondly, reduce formulas $S_2$ and $S_3$ to (\ref{S0}), they are expressed through cross-correlation in \cite{fitch2005fast}. Therefore $S_2$ and $S_3$ can be approximated with any desired precision by a sum of a few cross-correlations of functions which are obtained from
$f$ and $g$ by simple procedures in form $f_p=\cos(c_p f)$ or $f_p=\sin(c_p f)$ 
(where $c_p\in \mathbf{R}$). 

It is concluded that known standard methods for investigating similarity of images can be presented in the form of correlation (\ref{S}) or in a sum of a few ($P\in \mathbf{N}$) such correlations, that is in the form 
\begin{equation} S_{\mbox{standard}}(s) =\sum_{p=1}^{P} \sum_x f_p(x-s)g_p(x),   \label{standard}      \end{equation}
where $f_p$ and $g_p$ are modified images obtained from the initial $f$ and $g$ by applying an operation 
$\mathcal{O}_p$, and each operation $\mathcal{O}_p$ is shift-invariant, that is it commutes with an arbitrary image displacement. 

\subsection{Drawback of standard methods and an idea of focusing: a heuristic consideration}
The observation (\ref{standard}) allows critically judge all the considered method in a unified scheme. We are going to demonstrate a drawback of (\ref{S0}), and the rest of the described methods inherit this drawback from formulas (\ref{S}) and (\ref{standard}). The drawback is that it has over-sensitive reaction to geometric distortion of the ship in the two images, and this does not suit well our purposes. Rather small rotations of the ship will make the sought correspondence undetectable. This effect also is shown in this paper in experiments.

\begin{figure}
\centering
\includegraphics[width=3.4in]{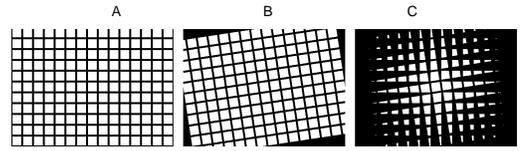}
 %where an .eps filename suffix will be assumed under latex, and a .pdf suffix will be assumed for pdflatex
\caption{Correlation of rotated images. (A) squares; (B) squares after rotation; (C) each square matches with its rotated version. }
\label{pic-rot}
\end{figure}

The drawback is presented in a heuristic form. It starts from a general observation that an actual image changes its values gradually from pixel to pixel, at least in most of its parts, and at some distance between two pixels these values become independent. We accept a simplification assuming that the image consists of small squares of constant values and these values are values of independent random variables. This is an approximation to a real image illustrated in Fig.~\ref{pic-rot}. Suppose we have such an image $f$ and its rotated version $g$. Consider a square grid of the introduced small squares covering the image $f$, they are shown in Fig.~\ref{pic-rot}(A) in white color. After rotation, these squares are changed to those in Fig.~\ref{pic-rot}(B).

We will examine how each square in Fig.~\ref{pic-rot}(A) matches with its rotated version in Fig.~\ref{pic-rot}(B). The matched areas are depicted in Fig.~\ref{pic-rot}(C). It can be seen from the Fig.~\ref{pic-rot}(C) that the number of the matched squares does depend on the angle of rotation only, and it is independent of size of the squares. The correlation (more precisely, its mathematical expectation) of $f$ and $g$ is proportional to the white area in Fig.~\ref{pic-rot}(C) divided by an area of one square.

From this heuristic construction, we draw the following conclusions:
\begin{enumerate}[(I)]
\item Described matching methods are expected to be rather sensitive to rotation of a ship. It can be seen from Fig.~\ref{pic-rot}(C), where rotation significantly diminishes number of white parts. 
\item Improving resolution of images will not improve detection of of the ship. This is illustrated by the fact that number of white parts in Fig.~\ref{pic-rot}(C) is fixed.
\item The matching can be improved if square areas change its size, smaller in the center, and gradually becoming bigger to the fringe. This can be interpreted as making the image artificially  smoother when farther from the center, and then using such an unevenly smoothed image for further correlation. 
While applying this idea to the whole image, we call it ``focusing", because a chosen part is well focused, and the other parts are out of focus, as shown in Fig.~\ref{Lena-eye}.
\end{enumerate}
The property (I) is widely presumed. For example, in  \cite{tsai2011shift} it is said that correlation methods are too sensitive in applications due to  ``distortion of the object surface under test".  In applications usually a small window is used, like in \cite{nielsen2011vision}, to make distortions less noticeable.  In our opinion, property (I) is the reason why the general problem of image registration is not yet solved satisfactory, and, instead of relying on machine vision (which necessitates perfect registration), other techniques are developed,  \cite{ park2011beacon}.

The effect (II) was empirically discovered in other circumstances such as  \cite{urhan2006modified}, where it was soundly proved in experiments that phase correlation method paradoxically benefits from down-sampling of the images when it concerns robustness to affine distortions.  

The idea (III) is widely used for small neighbourhoods of feature points as an empirical technique for articulating a feature point. The purpose is to make its neighbourhood more resilient to small rotations \cite{brox2011large}, and therefore it formally belongs to 
feature-based approach of registration techniques \cite{zitova2003image}.
 In the next section we modify this idea for area-based techniques.

\subsection{Focused correlation: a way of interpretation of spatial information}

\begin{figure}
\centering
\includegraphics[width=3.4in]{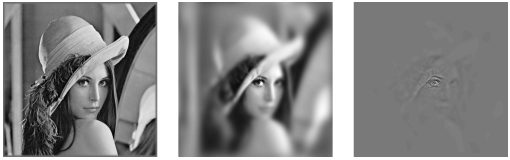}%{Lena_eye.eps}
\caption{{\it Left:} Initial crisp image; {\it Center:} Result of focusing with parameters: $\varepsilon=0.06$; focus $p_*$ is chosen in the center of an eye. Intuitively, it is obvious that such an image, being rotated for small angle, would coincide with the itself better than a crisp image would do, and therefore the image information is present here in a way that suits better for image registration in case of small rotation or, more generally, linear distortion with the fixed point at the focus; {\it Right:}  Result of focusing of the whitened image. }
\label{Lena-eye}
\end{figure}

We define focusing procedure $\mathcal{F}$ with parameter $\varepsilon>0$ and focus $p_*$, which is a position of a pixel. The parameter $\varepsilon$ determines strength of the focusing. 
For each point in the plane $x$ set a value $\sigma = \varepsilon \|x-p_*\|$, then the definition is  
\begin{equation}(\mathcal{F}(f))(x) \equiv \int_{\mathbf{R}^2}
      \frac{1}{2\pi \sigma^2} e^{-\frac{\|y-x\|^2}{2\sigma^2}} f(y)dy. \label{Gaussian}
\end{equation} 
The illustration follows on an example of the standard image "Lena" in Fig.~\ref{Lena-eye}. 
However, note that in the presented method we do not apply the focusing directly to the image, but for its whitened (randomised) version, because then we have controlled blurring, that is we know in which degree the image is blurred in its different parts. If we would apply the focusing to the initial image,   the resulting variable smoothness would not be known since the initial image already has  different unknown smoothness degree in different its parts. Moreover, our research demonstrates that application of focusing to the initial image produces rather negligible benefit, and, as we can suppose, this is the reason why the idea of \cite{brox2011large} was used locally only.    

%Another advantage of whitening is that it allows theoretical investigation of whitened images as random fields; and therefore a solid theory of the proposed method can be developed. However, we do not present it in this paper due to inevitably lengthy considerations.    

In short, we introduce an artificial controlled sensor measurement uncertainty for purpose to cope with really happening uncertainty of unknown geometric distortion. 
 
We define a focused correlation as a cross-correlation of images 
$\mathcal{F} f_1$ and $g_1$, where $f_1=\mathcal{O}(f)$ and $g_1=\mathcal{O}(g)$. Two cases are considered in the paper: 
\begin{itemize}
\item  $\mathcal{O}$ is the orientation operator defined above, then we have "Focused orientation correlation";  
\item  $\mathcal{O}$ is phase retaining operation defined above, then we have "Focused phase correlation".  
\end{itemize}
The operation $\mathcal{F}$ is not shift-invariant, therefore focused correlation differs from (\ref{standard}), and thus we genuinely present a new method.   

Intuitively it is apprehensible that the bigger the parameter $\varepsilon$, the bigger distortion the method can tolerate, however it is at expense of losing overall reliability since for bigger parameter $varepsilon$ information is lost due to smoothing. Therefore, a trade-off necessitates, and we use in our experiments $varepsilon=0.06$ which is defined empirically.  

According to our task, we  use two images and find a displacement which retains the most unchanged mutual information in them, and this should be a displacement of the ship since nothing else is expected in the open sea. The method in some sense is opposite to detection of vehicles in the land 
and cannot use an advantage of an unchanged background \cite{kafai2012dynamic}, 
\cite{chen2011hierarchical}. 
Focused correlation method correlates minuscule features in the image, and those change in the wave pattern and do not change in the ship pattern. On the contrary, coarser patterns, like a wake (i.e. long waves or a track left by a vessel) may remain stable. This is why we focus on fine structures, and it follows that we would prefer higher resolution in images, less compression in image information, and better randomised images, -- all these underline minuscule patterns. Another property of the focused correlation is that it is rather more robust, in  comparison with ordinary correlation, to small rotation and alike geometric distortion which the ship can undergo. All these properties are demonstrated in the presented experiments as follows. 

\section{Sea and ship separation in matching surface}\label{separ1}

\begin{figure}
\centering
\includegraphics[width=3.4in]{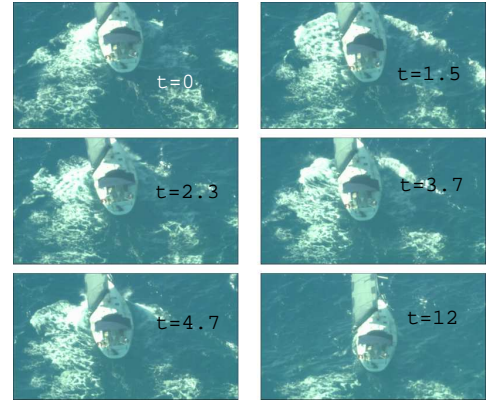}
 %where an .eps filename suffix will be assumed under latex, and a .pdf suffix will be assumed for pdflatex
\caption{A video scene taken during 12 seconds. The boat rocks, and at times $t=2.3$ and $t=4.7$ it is less rotated than at times $t=1.5$ and $t=3.7$. {\it By courtesy of SAGEM.}}
\label{SIXframes}
\end{figure}

To solve the task posed in Section \ref{Task}, we have to determine how to let the water under the UAV be gathered into one place, so that a dry ship may appear. Firstly we do this not in the initial real images, but in the matching surface; for the initial images the separation is introduced later in Section \ref{ImSepar}. Each point $s$ in the matching surface $S$ expresses the shift $s$ between the two initial images $f$ and $g$. However, the shift 
$s$ is meaningful, that is an area in the image with such shift exists, only if the value $S(s)$ is a few times above the standard deviation of all the values of $S$. 
To demonstrate  this idea, consider images of a boat in Fig.~\ref{SIXframes}. The first image is an initial image taken at time $t=0$ sec, and the rest of images were taken at times $t=$1.5, 2.3, 3.7. 4.7 and 12 seconds.         

\begin{figure}
\centering
\includegraphics[width=3.3in]{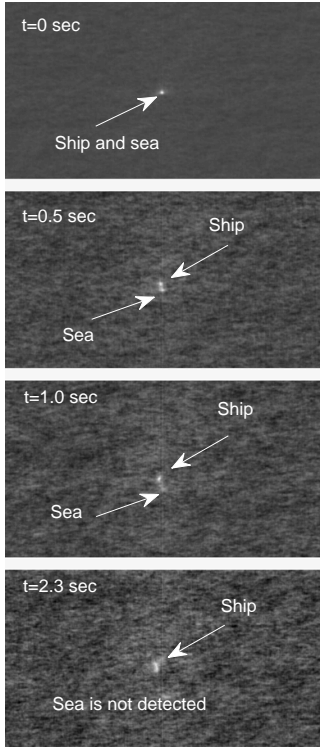}
\caption{Matching surface at different times corresponding video at Fig.~\ref{SIXframes}. The shift vector of the boat and the shift vector of the sea gradually separate, then the shift vector of the sea disappears, while the shift vector of the ship becomes presented by a blob due to rotation of the ship.}
\label{SEAvsSHIP}
\end{figure}

We scrutinize the initial period from $t=0$ to $t=2.3$ sec. in Fig.~\ref{SEAvsSHIP}. 
We  show the process of sea and ship separation. At the starting moment $t=0$ the sea and the ship both have zero displacement yet. This is reflected in the first image in  Fig.~\ref{SEAvsSHIP}. After $0.5$ sec. they have different motions: the sea pattern moves down quicker that the ship. In this, the second image of Fig.~\ref{SEAvsSHIP}, one can see that shift vector of the sea starts to loose its certainty, because different parts of the sea move differently. In the next image, $t=1.0$, this effect  is even more prominent. Eventually, in the last image of the matching surface the sea shift vector disappeared. The shift vector of the ship also suffered: it is not concentrated as before, because the ship changed it geometric appearance. 
From this experiment we conclude, that in about a second, the sea vanished from the matching surface, but the ship is still present.    

\begin{figure}
\centering
\includegraphics[width=3.3in]{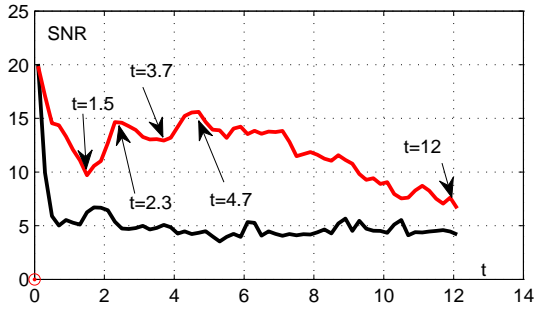}
\caption{SNR of the matching surface corresponding to the video in Fig.~\ref{SIXframes}. The used method is focused orientation correlation. Compare with caption of Fig.~\ref{SIXframes} for explanation of moments $t=1.5,\,2.3,\,3.7,\,4.7$.}
\label{SIXFocOC}
\end{figure}

This experiment is also illustrated in Fig.~\ref{SIXFocOC}, where two graphs are presented. 
The both graphs present signal-to-noise ratio (SNR) which is defined as a ratio of a maximum value of the matching surface to the standard deviation of the values of the surface, that is 
\begin{equation}
\mbox{SNR}(S) = \frac{\max(S)}{\operatorname{stdev}(S)}. \label{SNRdef} 
\end{equation} 
The upper graph is SNR of the matching surface between the initial frame at $t=0$ and a frame at $t>0$ from the video sequence partly shown in Fig.~\ref{SIXframes}. It is seen that when the ship's appearance in the images differ less, then SNR is higher. Since the ship rocks, the graph has a periodic appearance. The lowest SNR occurred at the moment $t=12$ sec. The latter demonstrates the limit of the method, and then for this particular image at $t=12$ sec. we estimate a geometric transform between the images as follows. Ship's rotation (comparatively with the time $t=0$) is 0.21 radian, and the scale along its length is 0.88. More precise description of the happened distortion of the ship from time $t=0$ to $t=12$ is given by an affine transform, which we estimated as  
$$      C= \begin{pmatrix} 0.92 & -0.20 \\
                           0.08  &  0.87
                           \end{pmatrix}   $$
The strength of the distortion in terms of norm is                            
\begin{equation}\| C-I \| =0.24, \,\, (I \mbox{ is 
an identical matrix}). \label{C-I}
\end{equation}

The second, lower, graph in the Fig.~\ref{SIXFocOC}, presents SNR of the the same images but without the boat. To eliminate the boat form the images, we just retained the left one third of the images, see Fig.~\ref{SIXframes}, and cut out the rest two third of it.  
This graph allows us to extract two bounding characteristics  $t_\text{sea}$ and SNR$_{\text{sea}}$ of the sea defined by their properties: 
\begin{itemize}
\item After passing a time interval of $t_\text{sea}$ sec presence of the sea vanishes in the matching surface.
\item The SNR of the sea (without boat) is bounded by  SNR$_{\text{sea}}$.
\end{itemize}

Now we can formulate results of this experiment.  From comparison of the two graphs in Fig.~\ref{SIXFocOC} we can conclude that $t_\text{sea}=1$ sec and SNR$_{\text{sea}}=7$. We also observed that the maximum detectable deviation of the ship is described by (\ref{C-I}). Therefore, presence of the boat in the image is indicated by conditions:
\[\text{SNR}> \text{SNR}_{\text{sea}} \, \text{ at a moment } t>t_\text{sea}.\] 
The value of $t_\text{sea}$ is empirical, while the value of SNR$_{\text{sea}}$ has some theoretical 
backing. Assuming that the matching surface is a Gaussian random field, we can estimate an expected value of SNR$_{\text{sea}}$ as $\sqrt{2\log N}$, see \cite{koval1997limit1}, where $N$ is the number of pixels in the matching surface. 
Taking SNR$_{\text{sea}}$ slightly bigger than that, we again come to the value SNR$_{\text{sea}}=7$.
    
\section{Comparison of the methods} \label{compar}
In the previous section we demonstrated how focused orientation correlation can 
benefit detecting a ship. In this section we compare different methods and come to a conclusion
that a combination of two methods is necessary.   
\subsection{Example of prevalence of Focused Phase correlation} 
\begin{figure}
\centering
\includegraphics[width=3.3in]{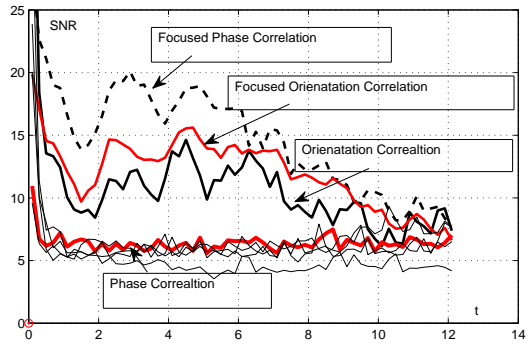}
\caption{Example of prevalence of focused correlation method. The graphs from Fig.~\ref{SIXFocOC} are present here for comparison of all the tried methods.} 
\label{SIXgraphs}
\end{figure}

\begin{figure}
\centering
\includegraphics[width=3.4in]{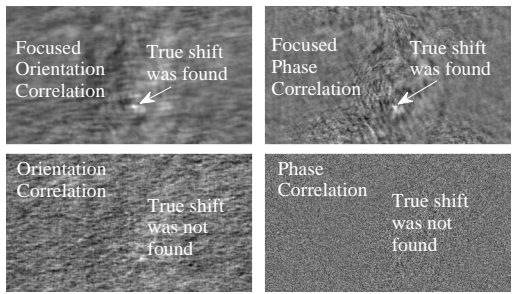}
\caption{Matching surfaces for the four methods for the last frame (at $t=12$) in Fig.~\ref{SIXframes}. 
Focused versions of the methods could detect the presence of the ship, while the original methods couldn't.}
\label{frame12sec.eps}
\end{figure}

Using the same video in Fig.~\ref{SIXframes}, we consider four methods: orientation correlation, phase correlation, and their focused versions. The results are present in 
Fig.~\ref{SIXgraphs}. 
Each of the methods produces two graphs, as shown in Fig.~\ref{SIXFocOC}, and the graphs from that figure are also presented in the Fig.~\ref{SIXgraphs}. The lowest four graphs are not labelled, they present SNR of the boatless left one third part of the scene. The greatest difficulty for ship detection is moment $t=12$ sec, and it is demonstrated separately by four matching surfaces in Fig.~\ref{frame12sec.eps}. From these data it is concluded that for this particular maritime scene the focused phase correlation method outperforms the rest.

\subsection{Example of prevalence of Focused Orientation correlation} 
\begin{figure}
\centering
\includegraphics[width=3.4in]{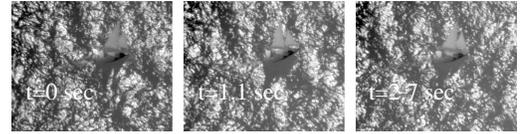}
\caption{A scene with lower frequency information in the ship area, and this leads to failure 
to detect presence/absence of a boat while using phase information for correlation; however, the gradient orientation information (in the form of focused orientation correlation) suffices for detecting.  
  {\it By courtesy of SAGEM.}}
\label{SECONDship.eps}
\end{figure}

\begin{figure}
\centering
\includegraphics[width=3.4in]{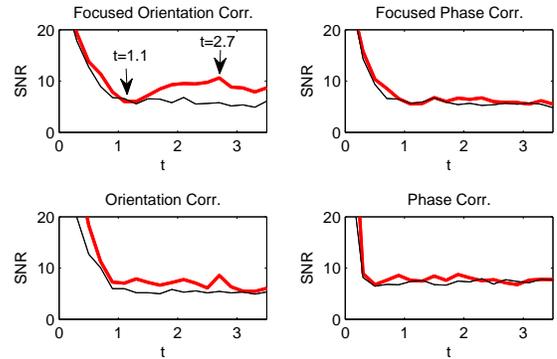}
 %where an .eps filename suffix will be assumed under latex, and a .pdf suffix will be assumed for pdflatex
\caption{Four methods applied to the scene in Fig.~\ref{SECONDship.eps}. The upper graphs (thick lines) present SNR of the matching surfaces of the whole picture. For comparison, the lower graphs (thin lines) present SNR for the left part of the picture which does not contain the boat.}
\label{FOURgraphs.eps}
\end{figure}

Consider an example of images with less hight frequency information, they are present in 
Fig.~\ref{SECONDship.eps}.
 For the four considered methods we have following graphs in Fig.~\ref{FOURgraphs.eps} arranged as before.  
These graphs show that in this case focused orientation correlation is the most reliable, while focused phase correlation is unable to provide a proper solution for ship detection.  

\subsection{Conclusion: Reliable method}
These two experiments illustrated our following findings: 
\begin{itemize}
\item Focused correlation can solve the task, while correlation itself only does not suit the task due to sensitivity to geometric distortion of the boat. We also tried other, simpler methods such as $S_0$, $S_1$, $S_2$, and found that they cannot provide a decent result for the scenes.   
\item For different kinds of scenes there is always one of the two variants of the focused correlation 
that works better, so we applied the both of them.
\item The boat was detected if one of the focused correlations robustly exceed value of SNR$_\text{sea}$ $=7$ at the time in between one and three seconds.
\item The maximum detectable deviation of the ship is described by (\ref{C-I}), however 
the reliable condition was found as $\| C-I \| <0.1$, (this corresponds to rotation less than $6^{\circ}$). The bound  $ t\le 3$ sec is an empirical value to guarantee that the ship normally rotates no more that to $6^{\circ}$. 
\end{itemize}  

These findings were confirmed in our experiments with more than 20 available maritime scenes.

\section{Sea and ship separation in images} \label{ImSepar}
To build an intelligent vision system we eventually will need the location of an object \cite{wangscalable}, that is to segment it in the image. Suppose a ship is detected, that is SNR $>$ SNR$_{\text{sea}}$ at time $t>t_\text{sea}$. The displacement between the ship images is found as a 2D vector $s_{\max}$ at which the matching surface 
reaches its maximal value (\ref{SNRdef}). If we align the second image, that is, obtain the shifted image 
 $g(x+s_{\max})$, then it should coincide with the first image $f(x)$ while $x$ belongs to the (unknown yet) ship area. With this observation we can determine the ship area.

\begin{figure}
\centering
\includegraphics[width=3.4in]{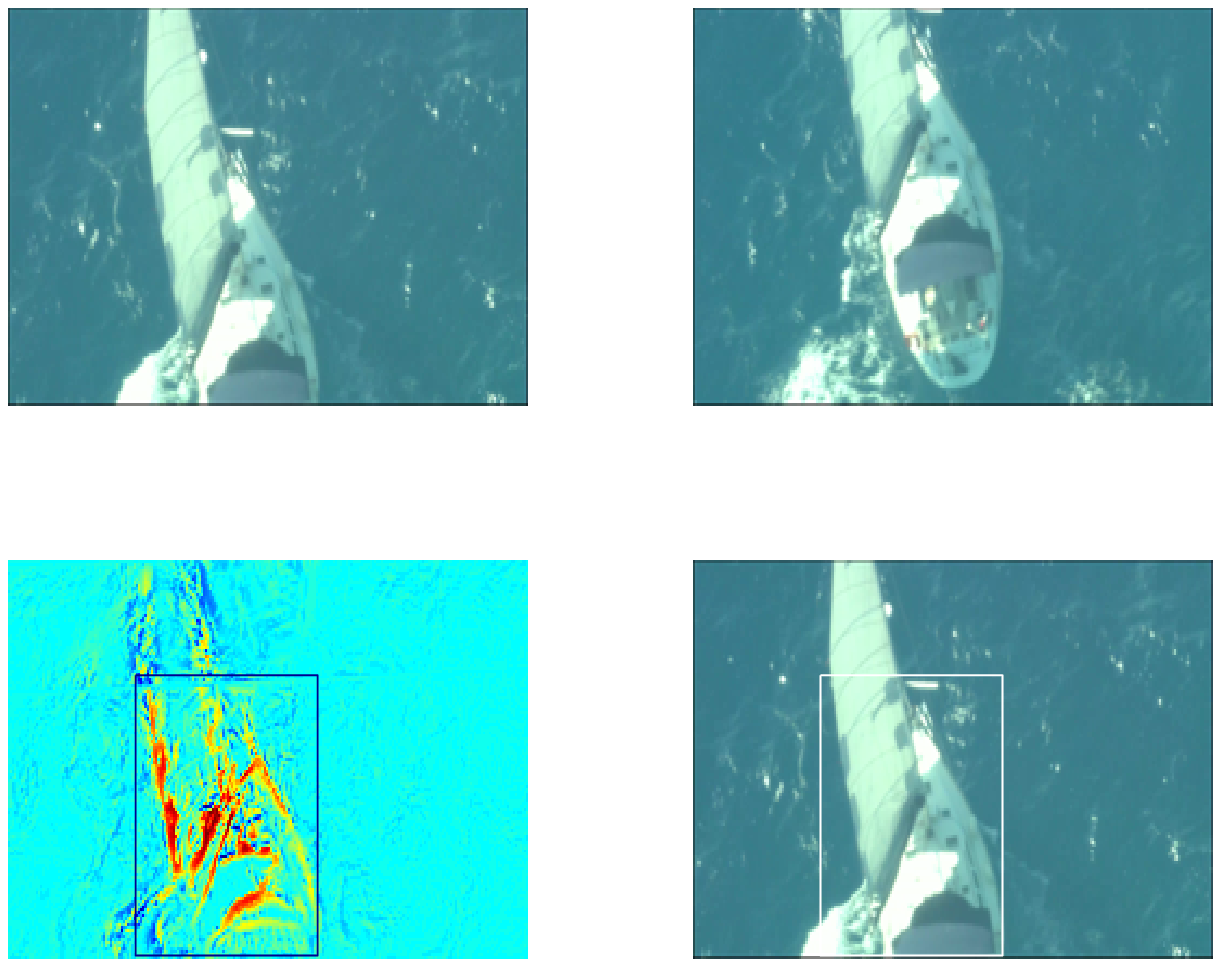}
\caption{Upper row: two pictures of a boat {\it by courtesy of SAGEM.} Lower row: matchability map and the segmented common area.}
\label{ShipSegm}
\end{figure}

\begin{figure}
\centering
\includegraphics[width=3.4in]{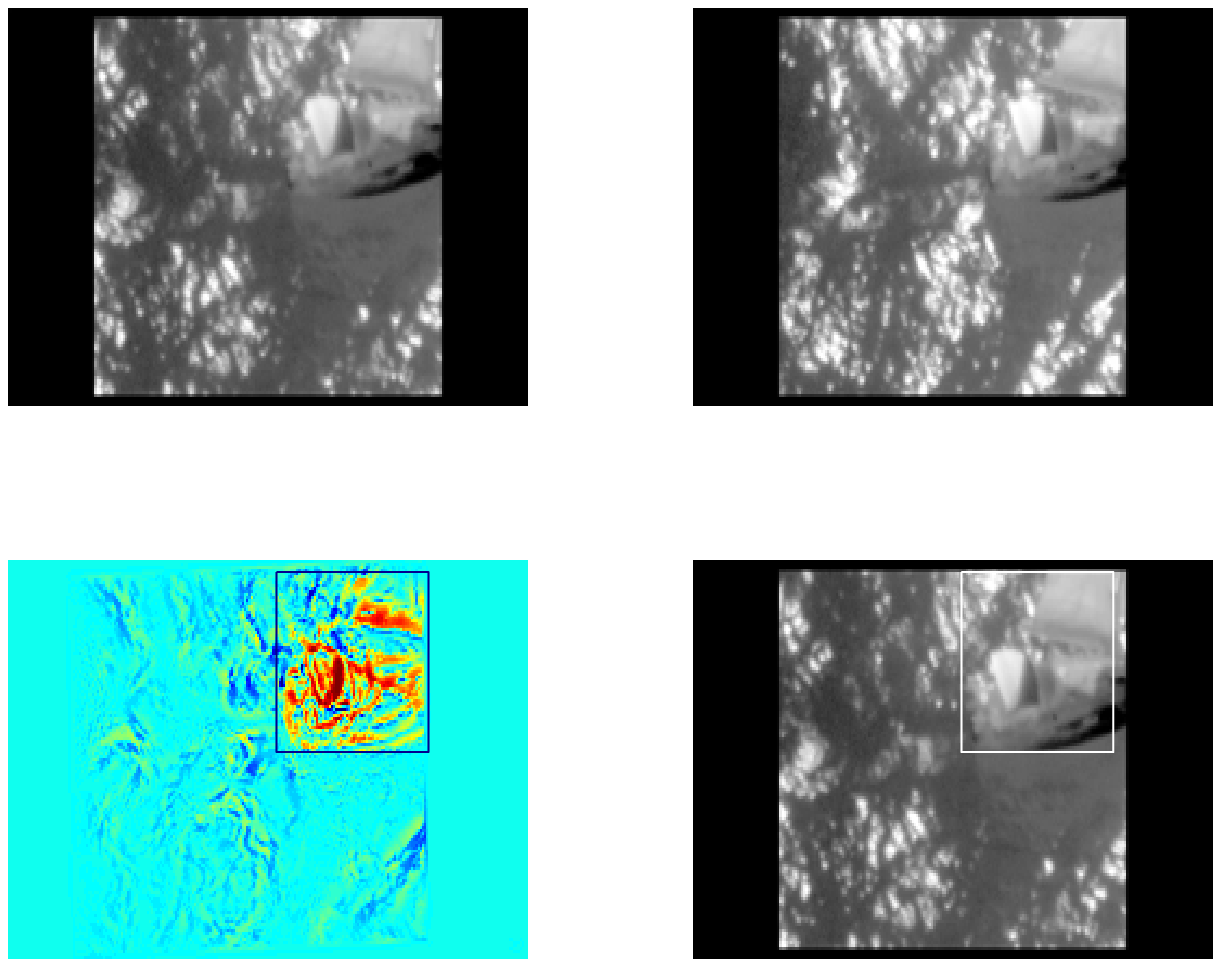}
\caption{Upper row: two pictures of a boat {\it by courtesy of SAGEM.} Lower row: matchability map and the segmented common area.}
\label{ShipSegm2}
\end{figure}

\begin{figure}
\centering
\includegraphics[width=3.4in]{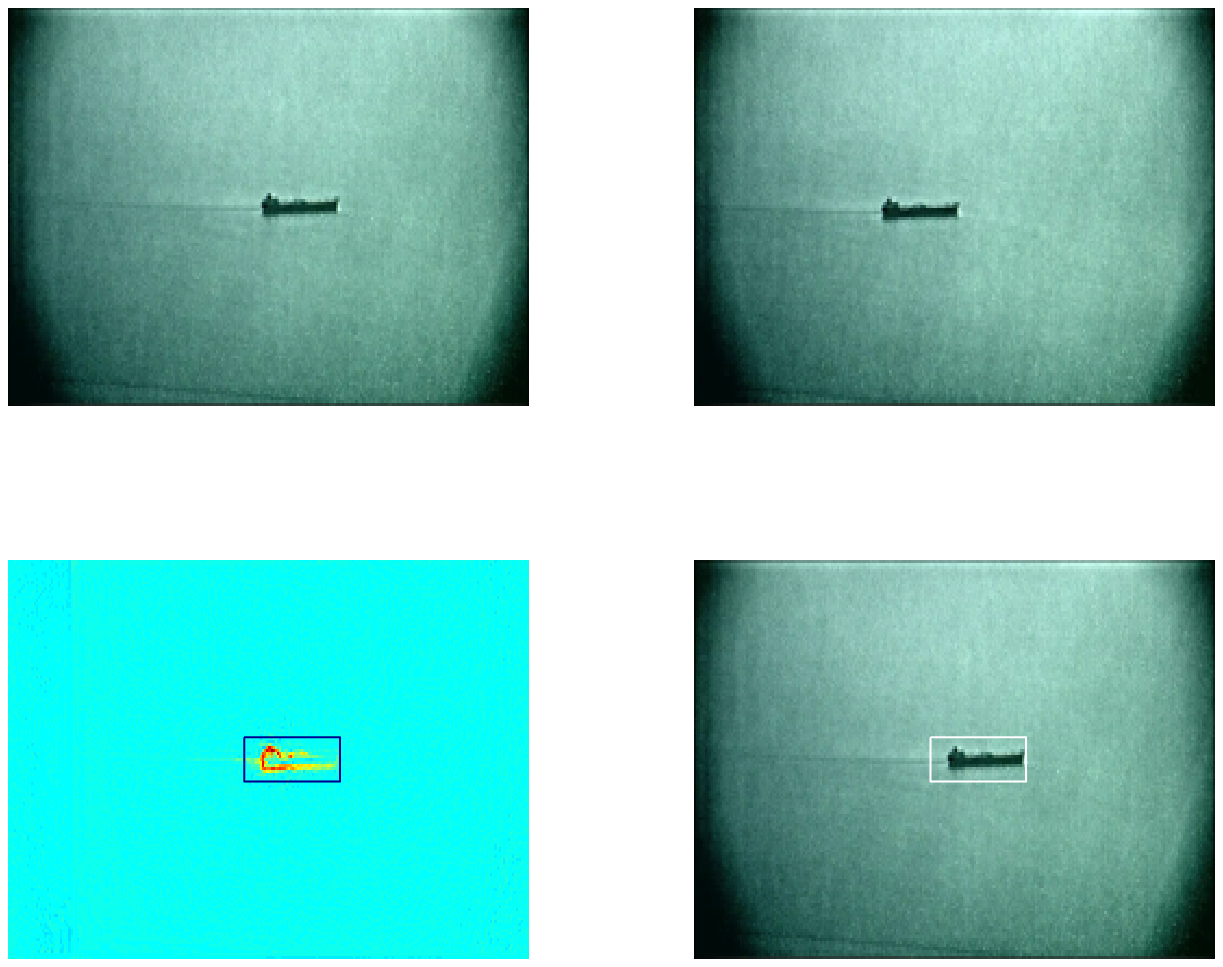}
\caption{Upper row: two pictures of a boat {\it by courtesy of SAGEM.} Lower row: matchability map and the segmented common area. }
\label{ShipSegm3}
\end{figure}

Our solution is to use the orientation operator $\mathcal{O}$, and compare images $\mathcal{O}f$ and $\mathcal{O}g(x+s_{\max})$. The ship area is then found as a set of points where 
the angle between  $\mathcal{O}f(x)$ and $\mathcal{O}g(x+s_{\max})$ is less than a particular 
number, which was empirically chosen to be $\arccos\, 0.4$. The result is visible on the matchability map which is defined as a cosine between unit vectors  $\mathcal{O}f(x)$ and  $\mathcal{O}g(x+s_{\max})$ at each pixel $x$. The matchability map shows which parts of the image can match its counterpart in the second image by displaying a degree of matching quality as a correlation coefficient ranging from $-1$ to $1$. 
The results of this automatic segmentation are present in Figs. \ref{ShipSegm}, \ref{ShipSegm2} and \ref{ShipSegm3}. These scenes present different degrees of difficulty for a human operator: obviously, for a human it would be the easiest to delineate the ship in Fig.~\ref{ShipSegm3}.

\section{Concluding Remarks}\label{discussion}
Due to long history of matching and correlating images, it seems rather difficult to propose a better  and feasible approach for applications. Papers on this topic appear at a rate of at least 100 papers each year, \cite{zitova2003image}. The same situation is with the ship detection topic, where many attempts were made, which seems to leave only a   possibility for incremental further development. 

We, however, proposed a novel way of substantial improvement of the most known method of correlation -- phase correlation.   This enables us to arrive at a new approach for boat detection and then resolve difficult cases when no prior information about statistical properties of the sea is available. The novelty presented in the paper can be therefore listed 
\begin{enumerate}
\item A new correlation method was proposed, and it shows reinforced robustness to geometric distortions of the involved images;
\item A new boat detection method was proposed, which is based entirely on comparison of images;
\item A usage of the observation that a fine pattern of the sea changes completely was proposed.  
\end{enumerate}   
  
The broader implication is that the proposed more robust modified phase correlation 
can be used everywhere where other similar correlation techniques are used. 
The proposed focused correlation benefit from higher resolution, therefore it can 
substitute the standard methods especially when larger images come about,   
for example medical images which usually have very high resolution. The main advantage is that the method needs no tuning and can even cope with scenes that are difficult to analyse for a human operator. 

We presented a novel correlation techniques that is able to align geometrically mutually translated and distorted pairs of 2D images. The method recovers the translational component of misalignment and it is more robust to small geometric distortion than known similar techniques. The method is based on a new way of interpreting spatial sensor information in the presence of geometric distortions.   

We examined the performance of the method in several maritime scenes compared to a few other methods as reference. The proposed method showed a low sensitivity to geometric distortion of the common areas and low sensitivity to surrounding changing background. In the considered maritime scene the common area is a ship (or boat) area, and the changing background is the image of the surrounding water in the sea. 
This enabled us to detect whether a ship is present in both images, since its presence manifests itself as an unchanged area that can be aligned. We also considered a direct extension of the method for further segmenting of the detected ship. This step proceeds by comparing aligned images. 

The method's behaviour was stable, which is promising for its usage for large variety of data. The detection of the ship was conducted by computation of movement between the two images only, and without  taking into consideration the image content as opposed to other methods. Therefore, the proposed method for ship detection can serve as a complement to the previously published work and then can be added as a new element to already working surveillance systems.

\section*{Acknowledgment}
The authors would like to thank SAGEM (SAFRAN) for providing the maritime videos. Our work is supported by SeaBILLA project funded under the 7th Research Framework Programme of the European Commission.
\bibliographystyle{IEEEtran}
\bibliography{shipBib.bib}
\end{document}